\documentclass[conference]{IEEEtran}
\IEEEoverridecommandlockouts
\usepackage{cite}
\usepackage{amsmath,amssymb,amsfonts}
\usepackage{mathtools}
\usepackage{algorithm, algorithmic}
\usepackage{graphicx}
\usepackage{textcomp}
\usepackage[numbers,sort&compress]{natbib}
\usepackage{caption}
\usepackage{balance}
\usepackage{cleveref}
\usepackage{color}
\usepackage{url}
\usepackage{float}
\usepackage{enumitem}
\usepackage{svg}
\usepackage{textcomp}
\usepackage{xcolor}
\usepackage{subcaption}
\usepackage{tabularx}
\usepackage{xurl, balance}
\usepackage{tabularray}
\usepackage{multirow}
\def\BibTeX{{\rm B\kern-.05em{\sc i\kern-.025em b}\kern-.08em
    T\kern-.1667em\lower.7ex\hbox{E}\kern-.125emX}}
\begin{document}

\title{Regression-Based Analysis of Multimodal Single-Cell Data Integration Strategies\\
}

\author{
\IEEEauthorblockN{Bhavya Mehta\IEEEauthorrefmark{1},
Nirmit Deliwala\IEEEauthorrefmark{2} and
Dr. Madhav Chandane\IEEEauthorrefmark{3}}
\IEEEauthorblockA{Department of Computer Engineering and Information Technology,
Veermata Jijabai Technology Institute\\
Mumbai, India.\\
Email: \IEEEauthorrefmark{1}bdmehta\_b19@ce.vjti.ac.in,
\IEEEauthorrefmark{2}nrdeliwala\_b19@it.vjti.ac.in,
\IEEEauthorrefmark{3}mmchandane@it.vjti.ac.in}}

\maketitle
\begin{abstract}
Multimodal single-cell technologies enable the simultaneous collection of diverse data types from individual cells, enhancing our understanding of cellular states. However, the integration of these datatypes and modeling the interrelationships between modalities presents substantial computational and analytical challenges in disease biomarker detection and drug discovery. Established practices rely on isolated methodologies to investigate individual molecular aspects separately, often resulting in inaccurate analyses.
To address these obstacles, distinct Machine Learning Techniques are leveraged, each of its own kind to model the co-variation of DNA to RNA, and finally to surface proteins in single cells during hematopoietic stem cell development, which simplifies understanding of underlying cellular mechanisms and immune responses. Experiments conducted on a curated subset of a 300,000-cell time course dataset, highlights the exceptional performance of Echo State Networks, boasting a remarkable state-of-the-art correlation score of 0.94 and 0.895 on Multi-omic and CiteSeq datasets. Beyond the confines of this study, these findings hold promise for advancing comprehension of cellular differentiation and function, leveraging the potential of Machine Learning.



\end{abstract}


\section{Introduction}

Omics relates to a comprehensive study of a large number of biological components in cells and organisms, including Genomics\footnote{\url{https://en.wikipedia.org/wiki/Genomics}}, Transcriptomics\footnote{\url{https://en.wikipedia.org/wiki/Transcriptomics}}\cite{transc}, Epigenomics\footnote{\url{https://en.wikipedia.org/wiki/Epigenomics}} \cite{epi}, Proteomics\footnote{\url{https://en.wikipedia.org/wiki/Proteomics}}\cite{pro} etc. Integrating these heterogenous datatyps from individual cells relates to the concept of "Multimodal Single Cell Data Integration (MSCDI)." This approach combines different layers of molecular information at the single-cell level, enabling a deeper understanding of cellular diversity, regulatory networks, and functional characteristics within complex biological systems. This integration is a powerful tool in modern biology, helping researchers uncover hidden patterns and relationships in single-cell data, leading to insights into various biological processes and disease mechanisms.

\begin{figure}[t]
    \centering
    \includegraphics[width=3in]{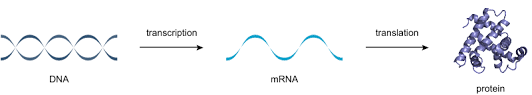}
    \caption{The flow of genetic information in a cell involves the transcription of DNA into mRNA, followed by translation at ribosomes to synthesize proteins.}
    \label{fig:intro}
\end{figure}

In the discovery of disease biomarkers, modelling the flow of materials from DNA to RNA and to Surface Proteins as shown in Fig.\ref{fig:intro}, helps researchers to pinpoint specific cell types or sub-populations that are implicated in diseases. Biomarkers derived from a single data modality may lack the specificity needed for accurate disease diagnosis and prognosis, as diseases exhibit cellular and molecular heterogeneity, even within the same individual. Multimodal integration enables the identification of distinct cell populations and their specific molecular signatures, offering insights into disease subtypes and progression. By analyzing gene expression, epigenetic marks, and other factors, researchers can identify cells with disregulated molecular profiles associated with the disease of interest signifying the importance of methods needed for such integration.

A lot of conventional techniques, although being currently used introduce limitations imposing logistical complexities and delayed results. While stacking of data layers may lead to high-dimensional dataset, making subsequent analyses challenging, Canonical Correlation Analysis assumes that the relationships between modalities are strictly linear. Moreover even techniques like Manifold Alignment and Probabilistic Models are computationally expensive, sensitive to noise and outliers in the data and may remove genuine biological variations that exist between batches. Even after these computations, researchers would often have to preprocess data manually by visually inspecting and clustering cells based on observed similarities in their characteristics, limiting scalability \& yielding data of variable quality. These challenges necessitate an integrated, data-driven approach. Machine learning (ML) algorithms provide a powerful solution to address these intricacies. ML leverages the computational prowess of computers to learn and identify patterns within multifaceted datasets. In the realm of MSCDI, ML presents an opportunity to merge and synthesize diverse molecular layers efficiently, enhancing our capacity to unravel the comprehensive cellular narrative.

In this paper, the aim is to overcome the shortcomings of traditional methodologies by  implementing Deep Neural Networks, Echo State Machines, Extreme Learning Machines and Random Forests to model the intricate data. The utilization of these models is driven by the fact that each of them employs a unique approach to model the data and train on regression tasks. This diversity allows us to gain a more profound understanding of the underlying principles that determine why one model outperforms another in specific scenarios which paves the way for guiding future research endeavors for MSCDI. The outcomes of our work hold the potential to accelerate innovation in methods for mapping genetic information across layers of cellular states by predicting one modality from another, pushing the boundaries of our knowledge in cellular biology.

\section{Related Works}
\citeauthor{Xiong2015} in \cite{Xiong2015} developed a machine-learning technique that scores how strongly genetic variants affect RNA splicing. More than 650,000 DNA variants were scored and the predictions also worked for outliers. \citeauthor{Alipanahi2015} in \cite{Alipanahi2015} uses DeepBind, a massively parallel deep-learning method used to predict the sequence specificities used for binding DNA with RNA. With an accuracy of 84.5\%, \citeauthor{Liu2010} in  \cite{Liu2010} makes use of random forest to identify the RNA binding sites in proteins. \citeauthor{Jansen2003} in \cite{Jansen2003} uses a Bayesian approach to integrate information protein-protein to thoroughly evaluate protein-protein interactions in yeast. Whereas \citeauthor{Tang2023} proposes UnitedNet, a deep neural network capable of combining various tasks such as multimodal integration and cross-modal analysis to analyze single-cell multimodality data.

SEURAT v3 by \citeauthor{Stuart2019} in \cite{Stuart2019}, merges scRNA data with ATAC-sequence to explore cell chromatin differences. LIGER in \cite{Welch2019} as proposed by \citeauthor{Welch2019}, integrates multiple cell datasets to create a low-dimensional space in which each cell is defined by one set of dataset-specific factors. \citeauthor{Korsunsky2019} in \cite{Korsunsky2019} uses Harmony to analyze a scalable join of scRNAseq across different modalities. MOFA \cite{Argelaguet2018} as proposed by\citeauthor{Argelaguet2018}, integrates datasets which helps map bilogical processes across various molecular tiers. ScAI is a method proposed by \citeauthor{Jin2020}, which helps analyze transcriptomic and epigenomic profiles within the same individual cells. Finally, \citeauthor{Cmero2021} in [11] introduces MINTIE, to identify not just fusion genes but all types of variants in transcriptomes.

\section{Multimodal Dataset}

The dataset encompasses a comprehensive collection of 300,000 multi-omics data points, each originating from individual cells acquired from four healthy human donors over a ten-day period, encompassing five distinct time points. The dataset is readily accessible through Link\footnote{\url{https://www.kaggle.com/competitions/open-problems-multimodal/data}}.

To cater to two distinct regression tasks, the dataset is partitioned into two subsets: "MultiSeq" and "CITEseq." The "Multiome" subset serves as the foundation for predicting gene expression (RNA) based on the chromatin accessibility (DNA) data, while the "CITEseq" subset is designed for quantifying surface protein levels using RNA-based predictions.

The MultiSeq dataset is characterized by a total of 105,942 samples, each endowed with 512 attributes. These attributes align with specific genomic locations, delineated by their respective genomic coordinates on the reference genome. They gauge the accessibility levels of the genome at these precise positions.

In contrast, the CITEseq dataset consists of 70,988 rows, each featuring 219 attributes—a configuration that mirrors the number of samples. Notably, the dataset is devoid of any missing values. These attributes encapsulate RNA expression levels for 22,050 genes, which have undergone a log1p transformation to preserve their non-negativity.

\section{Architectures}

\subsection{Deep Neural Networks}
Rather than relying on vanilla Deep Neural Networks, we introduce slight architectural modifications by integrating Self Multi Headed Attention (MHA) modules. 
The MHA mechanism, as proposed by \citeauthor{mha} in \cite{mha}, offers the capability to process distinct segments within an input sequence concurrently. This is in contrast to the sequential processing of the entire input sequence in a single step, as seen in Dense Layers. In the case of a 2D MHA, both the key and value inputs are projected onto each other, resulting in a unified output stream that demonstrates enhanced generalization and resilience to noise.

\begin{figure}[t]
    \centering
    \includegraphics[width=3in, height=2cm]{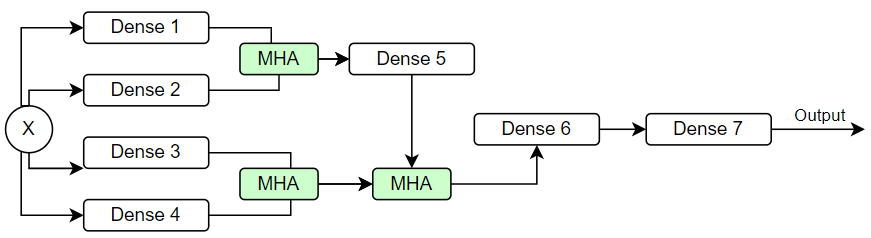}
    \caption{The model architecture combines various Dense layers with MHA modules with a total of 0.96M trainable parameters. Dense Layers 4, 5 and 6 have attached BatchNormalization layers as well to stabilize the gradient descent. A final Linear layer is used to give predictions for RNA and DNA data.}
    \label{fig:mha_arch}
\end{figure}

Our architectural design shown in Fig.\ref{fig:mha_arch},is robust, initially forwarding the input data through four distinct dense layers, each accompanied by Dropout and Relu Activation layers. The output of the first two layers is subsequently directed into separate MHA units. These MHA units merge the individual streams into a unified output, which is then further processed by a new MHA. It is essential to note that these MHA units function solely as operators and do not participate in the backpropagation process. Instead of merely concatenating data, these modules enhance feature extraction capabilities.

The final output from the MHA module is then channeled through Dense Layers, with a linear activation function, effectively serving as the output of the architecture.
\subsection{Echo State Network (ESN)}
\begin{figure}[t]
    \centering
    \includegraphics[width=3in]{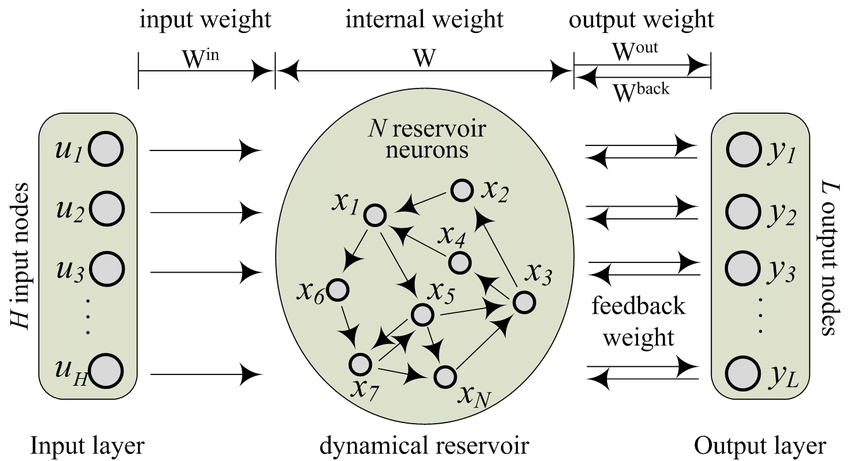}
    \caption{Overview of a general ESN architecture as proposed by \citeauthor{esn} in \cite{esn}. The input layer sends data to a pool of nodes which have feedback mechanisms with the output layer.}
    \label{fig:esn}
\end{figure}

ESNs are a reservoir computing technique that demonstrates considerable utility in the realm of multivariate regression. ESNs leverage a fixed, randomly generated reservoir of interconnected nodes to capture and amplify the temporal dependencies present in high-dimensional datasets  as shown in Fig.\ref{fig:esn}. These reservoir nodes are typically governed by a nonlinear activation function, facilitating the modeling of complex relationships among variables. The key advantage of ESNs lies in their ability to model intricate, nonlinear, and dynamic relationships among multiple variables as given by:
\begin{equation}
\begin{aligned}
    \mathbf{x}(t) &= \tanh(\mathbf{W}_{in} \cdot \mathbf{u}(t) + \mathbf{W} \cdot \mathbf{x}(t - 1)) \\
    \mathbf{y}(t) &= \mathbf{W}_{out} \cdot [\mathbf{x}(t), 1]
\end{aligned}
\end{equation}
Where, 
\begin{align*}
x(t) & : \text{State vector at time $t$ (internal state of ESN)} \\
{W}_{in} & : \text{Input-to-hidden weight matrix} \\
u(t) & : \text{Input vector at time $t$ (external input to ESN)} \\
{W} & : \text{Hidden-to-hidden weight matrix} \\
y(t) & : \text{Output vector at time $t$ (network's prediction)} \\
{W}_{out} & : \text{Output weight matrix (learned during training)} \\
\tanh & : \text{Hyperbolic tangent activation function} \\
[x(t), 1] & : \text{State vector extended with a bias term}
\end{align*}

\subsection{Extreme Learning Machines (ELM)}
\begin{figure}[t]
    \centering
    \includegraphics[width=2.4in, height=4cm]{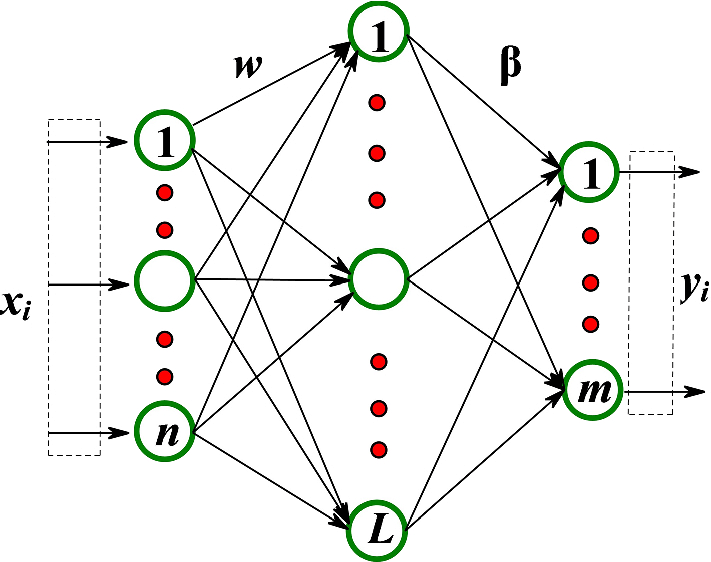}
    \caption{Overview of a general ELM architecture as proposed by \citeauthor{elm}\cite{elm}.}
    \label{fig:elm}
\end{figure}

ELMs excel in handling large-scale datasets and high-dimensional feature spaces, making them well-suited for complex regression problems. Their core advantage lies in their efficient training process, where hidden layer weights are randomly generated and then analytically determined using least-squares methods as shown in Fig.\ref{fig:elm}. Their random initialization and use of the entire training dataset during weight computation reduce the risk of overfitting. Furthermore, their fast training speeds and minimal hyperparameter tuning requirements make them a practical choice for real-world technical challenges where computational efficiency is paramount. The equation at each hidden node mainly follows: 

\begin{align}
H &= f(X \cdot W_{\text{in}} + b)  \\
W_{\text{out}} &= (H^+)^{-1} \cdot T \\
Y_{\text{new}} &= f(X_{\text{new}} \cdot W_{\text{in}} + b) \cdot W_{\text{out}}
\end{align}

Where,
\begin{align*}
H & : \text{Hidden layer output} \\
f() & : \text{Activation function used in the hidden layer} \\
X & : \text{Input data vector} \\
W_{\text{in}} & : \text{Input-to-hidden layer weight matrix} \\
{b} & : \text{Hidden layer bias vector} \\
{W}_{\text{out}} & : \text{Output weights for regression} \\
H^+ & : \text{Pseudoinverse of the hidden layer output matrix} \\
T & : \text{Target values for regression} \\
Y_{\text{new}} & : \text{Predicted output for a new input data point} \\
X_{\text{new}} & : \text{New input data point for prediction}
\end{align*}

\subsection{Random Forests (RF)}
\begin{figure}[t]
    \centering
    \includegraphics[width=3in]{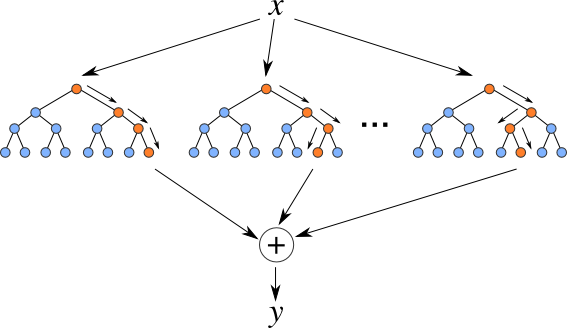}
    \caption{Random Forest as an ensemble of multiple decision trees.}
    \label{fig:rf}
\end{figure}

Random Forest\cite{rf} as given by \citeauthor{rf} is a robust ensemble machine learning algorithm with a strong technical foundation. It operates by creating an ensemble of decision trees as presented in Fig.\ref{fig:rf}, each of which is trained on a bootstrapped subset of the training data. RF introduces randomness by randomly selecting a subset of features for each decision split within a tree. The algorithm then combines the predictions of these trees through majority voting (for classification) or averaging (for regression) to enhance predictive accuracy and mitigate overfitting.

RF's technical strengths lie in its scalability, suitability for high-dimensional data, and robustness to noisy data and outliers.


\section{Implementation}
\begin{table*}
\centering
\caption{The table outlines a thorough comparison of the Correlation Score and Mean Squared Error (MSE) values for the mentioned methods on testing data.}
\label{tab:results}
\begin{tabularx}{\textwidth}{|c|*{4}{>{\centering\arraybackslash}X|}}
\hline
\textbf{Model} & \multicolumn{2}{c|}{\textbf{Gene Expression Prediction (Multiome)}} & \multicolumn{2}{c|}{\textbf{Surface Protein Level Prediction (CiteSeq)}} \\
\cline{2-3} \cline{4-5}
& Correlation Score & MSE & Correlation Score & MSE \\
\hline
DNN & 0.938 & 2.768 & 0.882 & 0.219 \\
\hline
ELM & 0.895 & 2.891 & 0.842 & 0.251 \\
\hline
\textbf{ESN} & \textbf{0.940} & \textbf{2.667} &\textbf{ 0.895} & \textbf{0.198} \\
\hline
Random Forest & 0.932 & 2.781 & 0.891 & 0.205 \\
\hline
\end{tabularx}
\end{table*}

\subsection{Data Preprocessing}
The data preprocessing pipeline begins with the removal of constant columns, reducing the initial count from 1194 to a more manageable set. The data is now converted to a sparse matrix because most of the data entries are zero, which are then projected to 512 dimension space by applying a Truncated Singular Value Decomposition (SVD), a memory-efficient implementation of the SVD \cite{svd}.
Subsequently, normalization is performed to ensure consistent scaling across features. After normalization, the Multi-ome subset retains all 512 columns where as Cite-seq subset retains 219 columns which contain the most significant features. This carefully refined dataset is then utilized as the input data for further analysis and modeling.


\subsection{Programming Environment}
The code is executed on a Kaggle P100 GPU infrastructure, a choice made to harness the CUDA libraries for accelerated computations. This GPU integration significantly enhances the execution speed and efficiency of the code, particularly when utilizing Keras for the implementation of machine learning models and related operations.

\subsection{Training Parameters}
Each and every model has been fine tuned and run on a platitude of combinations. The DNN model is run on a total of 300 epochs using Grouped K-Fold Validation Technique with a total of 3 splits. The ELM model was found to work best with a hidden layer size of 8000 neurons. For ESN, the MultiSeq data gave the best results with 512 reservoirs and spectral radius value of 0.4, while for CiteSeq data it was 100 and 0.4 respectively. This difference in parameters is mainly due to difference in dataset and attribute sizes. Finally, for the Random Forest Regressor,  max\_leaf\_nodes are set at 200 using all jobs in parallel.

\section{Results}
\begin{figure}[t]
    \centering
    \includegraphics[width=3in]{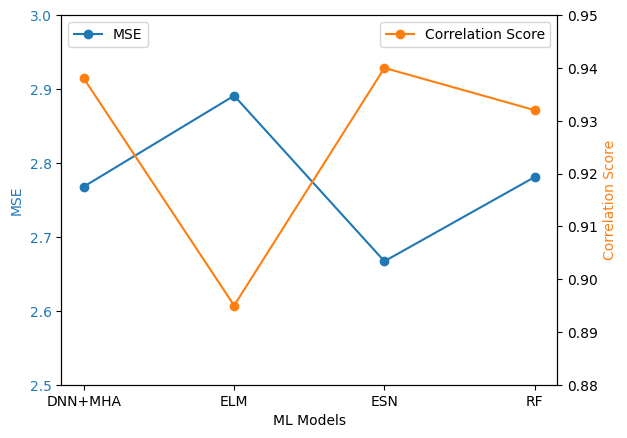}
    \caption{MSE and Correlation Score Comparision of models on MultiSeq Data}
    \label{fig:multi_result}
\end{figure}

\begin{figure}[t]
    \centering
    \includegraphics[width=3in]{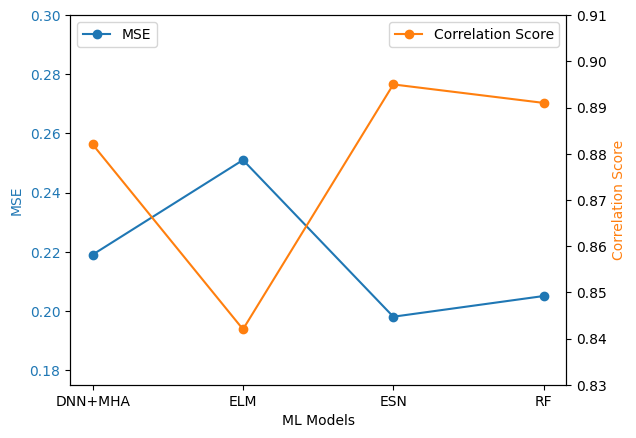}
    \caption{MSE and Correlation Score Comparision of models on CiteSeq Data}
    \label{fig:cite_result}
\end{figure}

The variation in performance among the machine learning models can be attributed to the specific characteristics of each model and the nature of the tasks involved. Table \ref{tab:results}
, Fig.\ref{fig:multi_result} and Fig.\ref{fig:cite_result} show a comprehensive comparison of each model's outputs.

\subsection{Gene Expression Prediction (Multiome):}
For this task, ESN achieved the highest Correlation Score of 0.940, which indicates strong linear association between predicted and actual values attributed to the model's architecture, well-suited for capturing the underlying relationships in the data.
Although DNN obtained a high Correlation Score of 0.938, its slightly higher MSE of 2.768 compared to ESN suggests that it introduced more error in its predictions, potentially due to the complexity of the model. Random Forests also have a good Correlation score of 0.932. ELM's higher MSE of 2.89, could be due to the model's limitations in capturing subtle relationships in the data owing to lack of control over hidden layer neurons and overfitting, especially when the number of hidden neurons is large.

\subsection{Surface Protein Level Prediction (CiteSeq):}
ESNs and Random Forest have numbers very close to each other. ESNs although with less number of reservoirs, excelled in this task with a Correlation Score of 0.895 and a low MSE of 0.198, showcasing its capability to accurately predict surface protein levels. Random Forests give similar results but take up even more memory to store the ensemble of decision trees. DNN performed well with a Correlation Score of 0.882 and a relatively low MSE of 0.219. ELMs again had an inferior performance as it may not be able to generalize well to tasks with significantly different data distributions.

\section{Conclusion}
In conclusion, this research endeavor explores and assess a diverse array of Machine Learning methodologies, each distinct from the other, with the primary goal of predicting gene expression and protein levels. Through experimentation and analysis, it becomes evident that although the DNN gives scores of 0.938 and 0.882 respectively, the ESN stands out as the top-performing model due to its remarkable adaptability to complex data patterns with a benchmark correlation score of 0.94 and 0.895. The paper thus represents exciting prospects for the continued development of predictive models in the realm of gene expression and protein level prediction, promising to unlock even greater accuracy and insight.

\section{Future Scope}
To propel this research further and enhance our predictive capabilities, researchers can consider more advanced modifications to the ESN framework. One promising avenue involves the integration of multi-layer reservoirs, allowing for hierarchical feature extraction and a more comprehensive understanding of the underlying data dynamics. Alternatively, instead of relying on entirely random connections, the incorporation of complex recurrent reservoirs, such as long short-term memory (LSTM) units or gated recurrent units (GRUs), could offer improved learning and modeling capabilities.

{
\scriptsize
\bibliographystyle{IEEEtranN}
\bibliography{bibliography}
}
\balance

\end{document}